\documentclass[pmlr]{jmlr}

\usepackage{longtable}

 \usepackage{booktabs}
 
 \usepackage{tablefootnote}
 \usepackage{mathtools}
 \usepackage[shortlabels]{enumitem}
 \usepackage{makecell}
 
 \DeclarePairedDelimiter{\ceil}{\lceil}{\rceil}
 \newcommand{\fatmidrule}{\specialrule{0.08em}{0.3em}{0.3em}}
 
\usepackage[load-configurations=version-1]{siunitx} 

\makeatletter
\def\set@curr@file#1{\def\@curr@file{#1}} 
\makeatother

\theorembodyfont{\upshape}
\theoremheaderfont{\scshape}
\theorempostheader{:}
\theoremsep{\newline}

\jmlrvolume{}
\jmlryear{}
\jmlrworkshop{}

\title[ExBEHRT]{ExBEHRT: Extended Transformer for Electronic Health Records to Predict Disease Subtypes \& Progressions}

\author{\Name{Maurice Rupp}
       \Email{maurice.rupp@gmail.com}\\ 
       \addr Novartis Oncology AG\\
       Basel, Switzerland
       \AND
       \Name{Oriane Peter}
       \Email{oriane.peter@gmail.com}\\ 
       \addr Novartis Oncology AG\\
       Basel, Switzerland
   		\AND
   		\Name{Thirupathi Pattipaka}
   		\Email{thirupathi.pattipaka@novartis.com}\\ 
   		\addr Novartis Oncology AG\\
   		Basel, Switzerland}

\begin{document}

\maketitle

\begin{abstract}
In this study, we introduce ExBEHRT, an extended version of BEHRT (BERT applied to electronic health records), and apply different algorithms to interpret its results. While BEHRT considers only diagnoses and patient age, we extend the feature space to several multimodal records, namely demographics, clinical characteristics, vital signs, smoking status, diagnoses, procedures, medications, and laboratory tests, by applying a novel method to unify the frequencies and temporal dimensions of the different features. We show that additional features significantly improve model performance for various downstream tasks in different diseases. To ensure robustness, we interpret model predictions using an adaptation of expected gradients, which has not been previously applied to transformers with EHR data and provides more granular interpretations than previous approaches such as feature and token importances. Furthermore, by clustering the model representations of oncology patients, we show that the model has an implicit understanding of the disease and is able to classify patients with the same cancer type into different risk groups. Given the additional features and interpretability, ExBEHRT can help make informed decisions about disease trajectories, diagnoses, and risk factors of various diseases.
\end{abstract}

\section{Introduction}
Over the past decade, electronic health records (EHRs) have become extremely popular for documenting a patient's medical history, with many existing records combining heterogeneous temporal information about diagnoses, procedures, laboratory tests, observations and demographic data from a variety of sources (primary care, hospital visits, etc.). In general, a sequence of medical events of a single patient is referred to as a \textit{patient journey}. Given the immense amount of data available (datasets range up to over 100M patients) and its level of detail, there is incredible potential for the use of machine learning to provide new insights into disease pattern recognition, early detection of rare diseases, and personalised risk prediction and treatment planning.\newline
Embedding algorithms derived from natural language processing (NLP) have shown remarkable performance when trained to represent patients' medical histories. Due to the chronological structure of EHRs, such algorithms can provide various insights into disease trajectories and clinical phenotypes. Recent advances in NLP have also shown that transformer-based methods such as BERT (\cite{Devlin2018}), GPT-3 (\cite{Brown2020}) and their variations are significantly superior to other approaches, as they are able to model complex temporal dependencies over a long period of time.
\newline
In this paper, we present a novel approach to incorporate multimodal features into Transformer models by adding medical concepts separately and vertically, rather than chaining all concepts horizontally. We show that these features are important in various downstream applications such as mortality prediction, patient subtyping and disease progression prediction.
\subsection*{Generalizable Insights about Machine Learning in the Context of Healthcare}
The main contributions from this work can be summarized as follows:
\begin{enumerate}
	\item A novel form of incorporating any sort of multi-modal EHR features into BERT (or any other Transformer-based model) without having to extend the resources needed to train the model due to consistent, fixed patient journey sequences.
	\item The addition of patient information that, to our knowledge, was not included in any previous work (BMI, smoking status, laboratory values) and improves model performance for several downstream tasks. These additional features provide a more comprehensive and complex understanding of patients, leading to deeper and more robust insights for clinicians when interpreting model results. In combination with the expected gradients model explainability, we can gain new insights into the different pieces of information and their impact on the outcome.
	\item An exploration of unsupervised clustering of cancer patients using the patient representation of ExBEHRT, identifying groups of cancer types and subgroups within one cancer type with diverse information about their characteristics for recognizing risk subtypes and treatment patterns.
\end{enumerate}

\section{Related Work}
Recent studies have adapted transformers to structured EHR data and shown their superiority in various benchmarks compared to other similar algorithms (\cite{Kalyan2022}). Since most publications in this area are a derivative of BERT (\cite{Devlin2018}), in this section we will focus exclusively on BERT-based approaches applied to EHR data.\newline
The first adaptation of EHR to BERT, called BEHRT (\cite{Li2020}), incorporated diagnosis codes and ages from EHRs and added additional embeddings to separate individual visits (segment embedding) and a position embedding for the visit number. To separate visits, the authors added \textbf{SEP} tokens\footnote{In NLP, tokens usually refer to the smallest unit into which an input is decomposed. This can be a word, part of a word or, as in this case, a medical concept}. between visits, analogous to the \textbf{SEP} token between sentences in BERT and \textbf{CLS} token as an artificial start token. The model was pre-trained by using the Masked Language Modelling (MLM) objective on diagnosis concepts.
\newline
Med-BERT (\cite{Rasmy2021}) introduced a code serialisation embedding in addition to diagnosis and position embeddings, indicating the order of diagnoses within a visit. Med-BERT was pre-trained with MLM and a binary classification target of whether a patient had at least one hospital stay of more than one week (\textit{prolonged length of stay in hospital} or PLOS). \newline
CEHR-BERT (\cite{Pang2021}) and BRLTM (\cite{Meng2021}) contain many more measures than the other two approaches. Instead of separate a diagnosis embedding, the studies combined all medical concepts (i.e. conditions, procedures and medications) of a patient into a single vector. This method results in considerable overhead when training a model, as the maximum length of the patient journey is significantly higher than if only the diagnosis codes were included. Adding more features (e.g. observations) would increase the resources required due to the increased length of the vector. \newline
In addition, there are a variety of models that either combine the BERT architecture with other machine learning models (\cite{Shang2019}, \cite{Poulain2022}, \cite{Li2021}) or focus exclusively on specific use cases (\cite{Azhir2022}, \cite{Prakash2021}, \cite{Rao2022}).
\newline
All the aforementioned approaches either lack generalizability to different domains due to specific pre-training (a key advantage of transfer learning using transformers), do not incorporate enough variety in patient information to generate informed decisions or are limited in the amount of data of a single patient they can process.

\section{ExBEHRT for EHR Representation Learning}
\begin{figure}[h!]
	\centering 
	\includegraphics[width=7.2cm]{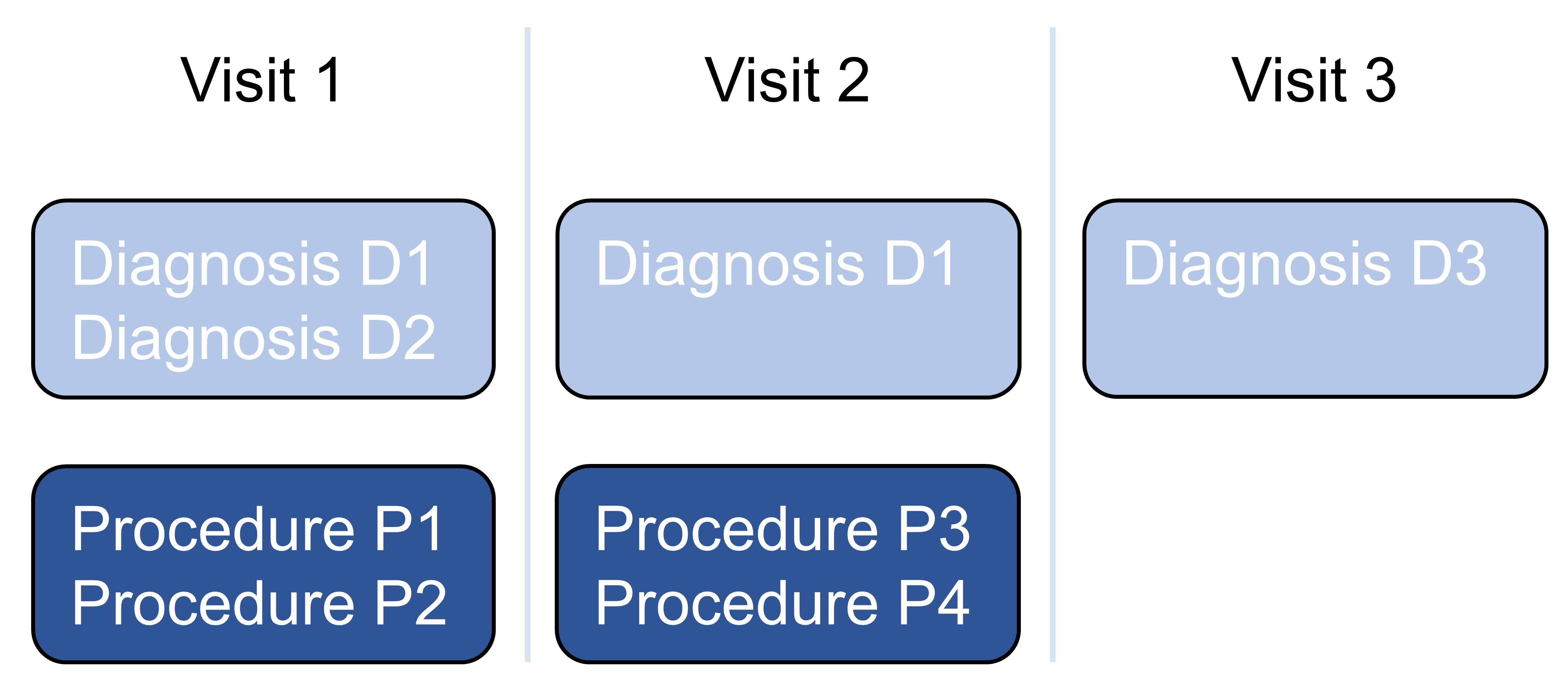} 
	\caption{An example of the procedures and diagnoses of a patient with three visits.}
	\label{fig:pt} 
\end{figure}

\begin{figure}[h!]
	\centering 
	\includegraphics[width=15cm]{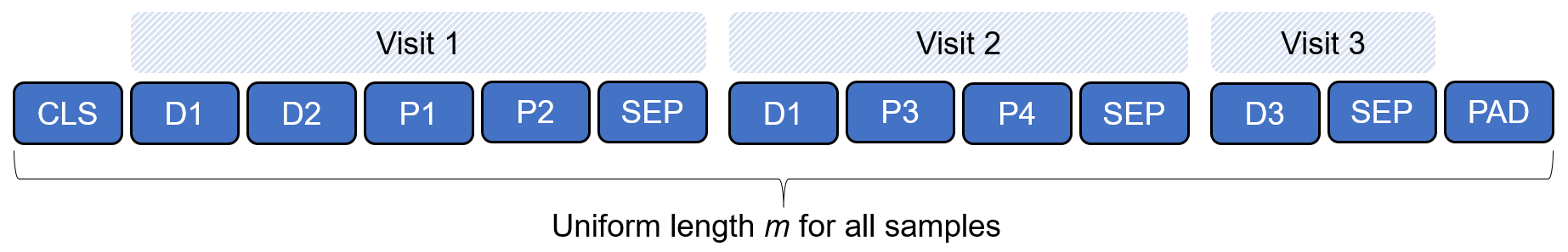} 
	\caption{An example of how models like CEHR-BERT and BRLTM represent the patient from figure \ref{fig:pt} by horizontally stacking all features into a 1D representation. Note that each additional measure potentially increases the maximal sentence length $m$.}
	\label{fig:brltm_repr} 
\end{figure}

\begin{figure}[h!]
	\centering 
	\includegraphics[width=13.5cm]{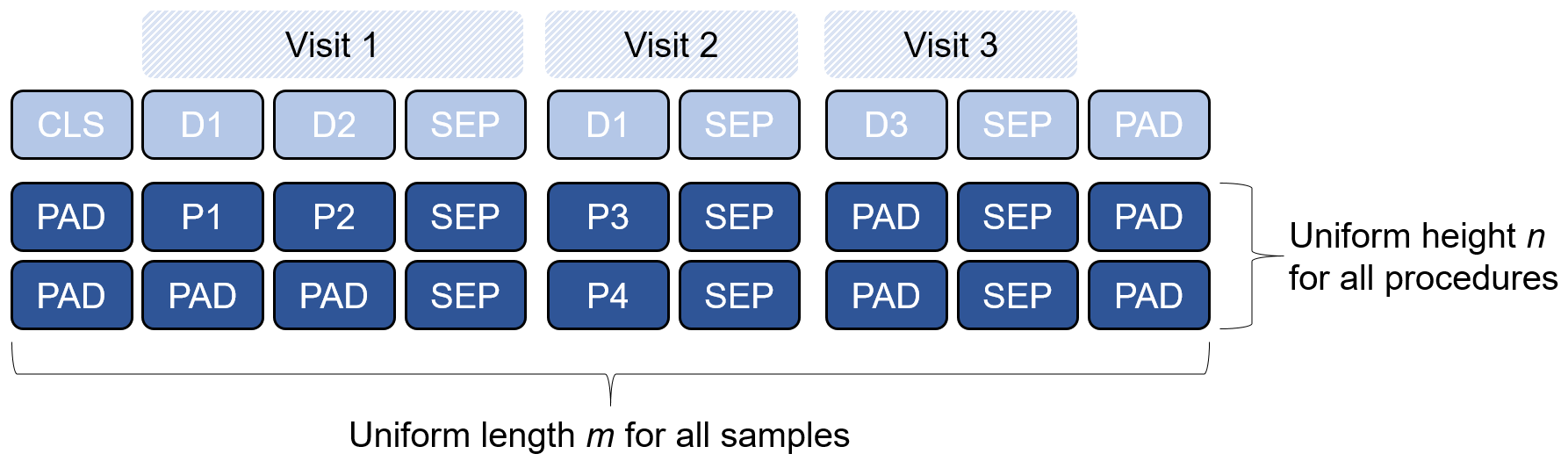} 
	\caption{An example of how ExBEHRT represents the patient from figure \ref{fig:pt}. As the features are stacked vertically, additional concepts (such as labs as shown in figure \ref{fig:embedding}) will not increase the sentence length $m$.}
	\label{fig:exbehrt_repr} 
\end{figure}

\begin{figure}[h!]
	\centering 
	\includegraphics[width=15.3cm]{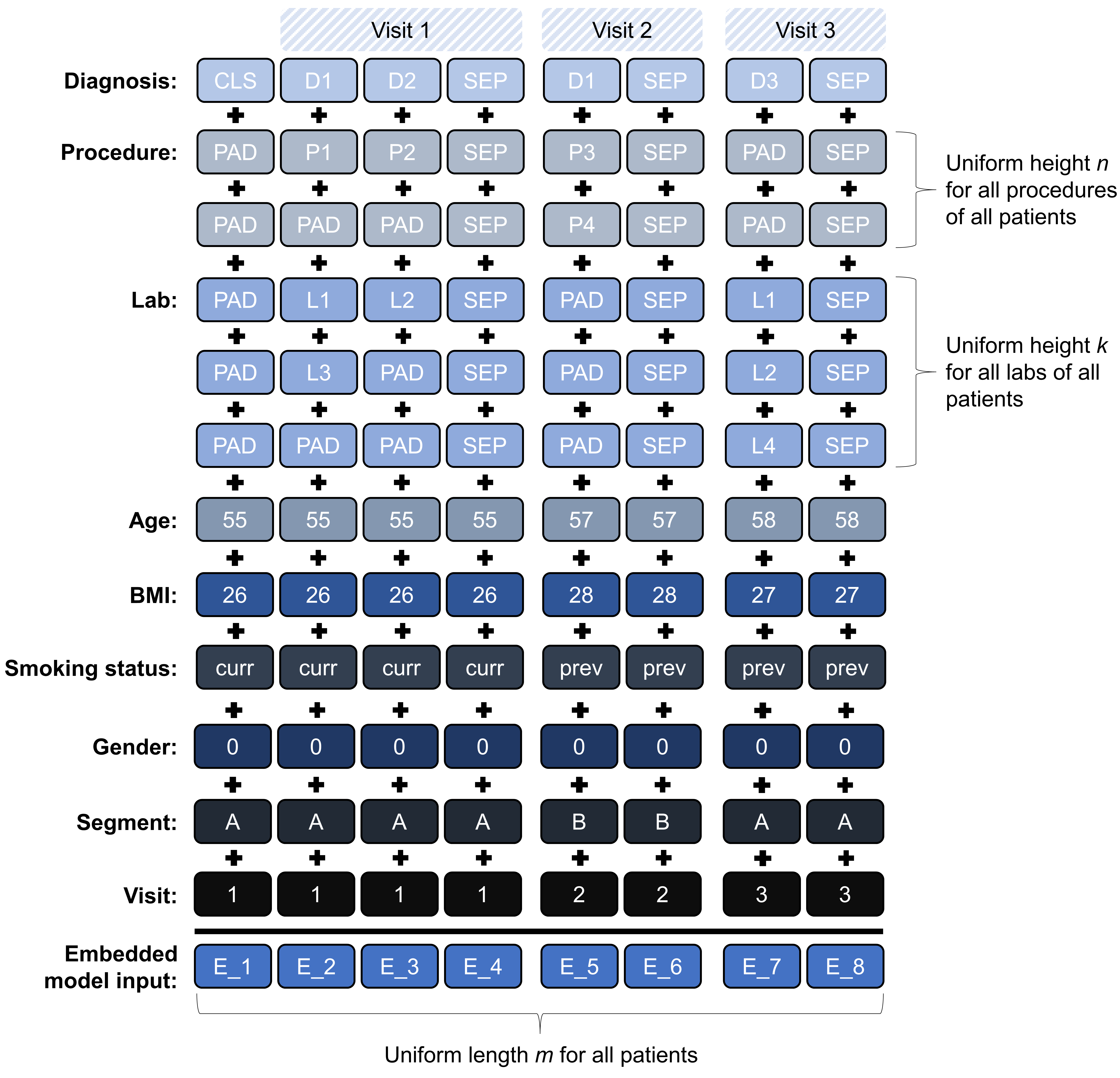} 
	\caption{A sample input of ExBEHRT. Each of the concepts has its own embedding, where each of the tokens is mapped to a 288-dimensional vector, which is learned during model training. After embedding, all concepts are summed vertically element-wise to create a single $288 \times m$ dimensional vector as input for the model.}
	\label{fig:embedding} 
\end{figure}

ExBEHRT is an extension of BEHRT where medical concepts are not concatenated into one long vector (as in Figure \ref{fig:brltm_repr} for the example patient shown in Figure \ref{fig:pt}), but grouped into separate, learnable embeddings per concept type. In this way, we avoid exploding input lengths when adding new medical features and give the model the opportunity to learn which concepts it should focus on. From a clinical perspective, it would also be stringent to separate diagnoses, procedures, drugs, etc., as they have different clinical value for downstream applications. We take the number of diagnoses in a visit as an indicator of how many "horizontal slots" are available for other concepts in that visit (e.g. two for the first visit in Figure \ref{fig:exbehrt_repr}). Therefore, the maximum length of the patient journey is defined by the number of diagnosis codes of a patient, regardless of the number of other concepts added to the model.
 \newline
As shown by the procedures in figure \ref{fig:exbehrt_repr}, but carried out in the same way with lab tests, there are three possible cases of adding a new concept to a visit:
\begin{enumerate}[a)]
	\item The number of procedures is equal to the amount of horizontal slots available in the visit (visit 1 - two each). The procedures can therefore be represented as a 1D vector.
	\item The number of procedures exceeds the amount of slots available in the visit (visit 2 - one diagnosis, two procedures). Here, the procedures fill up the number of horizontal slots line by line until there are no more procedures left, resulting in a 2D vector of dimensions $\#slots \times \ceil{\frac{\#procedures }{\#slots}}$.
	\item The number of procedures subceeds the amount of slots available (visit 3 - one diagnosis, no procedures). The procedures are represented as a 1D vector and then padded to the amount of horizontal slots available.
\end{enumerate}
The padding token \textbf{PAD} can be understood as an indicator to the model which parts of a patient journey can be neglected, as they don't contain information. It is added at the end of a sentence to ensure the same length $m$ for each patient. After the reshaping described above, all procedures of all patients are padded to the same amount of rows $n$ to enable batch processing. $n$ is set to the .95 percentile over all representations of visits of all patients before training. Since BMI, smoking status and gender naturally don't fluctuate within one visit, they do not need to be rearranged in a complex way. Therefore, the value recorded within a visit is copied to all horizontal slots of the corresponding visit. In addition, the embeddings of diagnoses, age and segment are the same as described in the original BEHRT publication. Before the inputs are passed to the model, each token is embedded in a 288-dimensional vector and all tokens are summed vertically. A visualisation of the complete input can be found in Figure \ref{fig:embedding}.

\subsection{Pre-training with Masked Language Modelling}
For pre-training the BERT-based models, we applied the standard MLM procedure described in the original BERT paper (\cite{Devlin2018}) applied on diagnosis code prediction using their BertAdam optimizer with cross-entropy loss. We followed the vast majority of subsequent papers, where in each iteration 15\% of the diagnosis codes of a patient are selected randomly and either masked (80\% of the time), replaced with another diagnosis code (10\% of the time) or kept the same (10\% of the time).
\subsection{Fine-tuning on Disease-Specific Event Prediction}
We validated our model on several disease-specific binary classification tasks from different domains. One oncology-specific task (\cite{Lu2022}) commonly found in the literature is the prediction of cancer patient mortality within six and twelve months. The observation window (the information provided to the model) is the entire patient journey to the first cancer diagnosis (including the visit with the first cancer diagnosis).\newline
The third task, which was part of the CEHR-BERT paper, is to predict the readmission of a patient with heart failure to the hospital within 30 days after heart failure. The observation window includes all visits within one year before the (first) heart failure.\newline
To account for the strong class imbalance between positive and negative outcomes, we also included focal loss (\cite{Lin2017}) in our hyperparameter search space. Focal loss reduces the relative loss for well-classified examples and puts more emphasis on difficult, misclassified examples.
\subsection{Cancer Patient Clustering with ExBEHRT Embeddings} \label{ssec:cancer_clust}
To generate patient clusters and visualize them in a meaningful way, we applied a combination of the dimensionality reduction technique UMAP (\cite{McInnes2018}) and the clustering algorithm HDBSCAN (\cite{Campello2013}). As ExBEHRT is not specialized on a specific disease, we conducted another pass of pre-training, where we initialized the model weights with the pre-trained ExBEHRT weights and applied MLM on cancer diagnosis codes only. This non-finetuned model was then used for generating the patient embeddings. After training, we conducted the following steps for the unsupervised clustering:
\begin{enumerate}[a)]
	\item Generate the ExBEHRT embedding (vector of size 288) for each patient (stemming from the CLS token at the beginning of each patient journey)
	\item Reduce the embeddings' size from $a)$ from 288 to 10 using UMAP to get representations for clustering and avoid the curse of dimensionality
	\item Cluster the 10-dimensional vectors using HDBSCAN
	\item Reduce the embeddings' size from $a)$ from 288 to 2 using UMAP to get 2D coordinates for each patient
	\item Visualize the Clusters from $c)$ in the 2D space of the embedding from $d)$
\end{enumerate}

\section{Cohort}
In this study, we used the Optum\textsuperscript{\tiny\textregistered} de-identified EHR database. It is derived from healthcare provider organizations in the United States, which include more than 57 contributing sources and 111,000 sites of care including hospital-based medical services networks comprising academic, private, and community hospitals treating more than 106 million patients. Optum\textsuperscript{\tiny\textregistered} data elements also include demographics, medications prescribed and administered, immunizations, allergies, lab results (including microbiology), vital signs and other observable measurements, clinical and hospitalisation administrative data, and coded diagnoses and procedures. The population in Optum\textsuperscript{\tiny\textregistered} EHR is geographically diverse, spanning all 50 US states.

\subsection{Pre-Training Cohort}\label{ssec:cohort}
We selected only data points collected during hospitalisations to ensure the quality and consistency of the data\footnote{This includes emergency patients, inpatients, observation patients, nursing facility patients, hospice patients and inpatient rehabilitation patients.}. Each patient must have at least five visits with valid ICD-9 or ICD-10 diagnosis codes to ensure sufficient temporal context.  Our cohort is selected with the same criteria as in BEHRT, resulting in 5.4M individuals. In order to prevent any sort of data leakage during pre-training and fine-tuning, the data is split into three datasets before training: training (80\%), validation (10\%) and testing (10\%).

\begin{table}[h!]
	\centering
	\caption{Statistics of the pre-training cohort.}
	\begin{tabular}{ ll } 
		\hline
		\toprule
		\textbf{Feature} & \textbf{Metric} \\
		\midrule
		Birth year &  1973$\pm$25, min: 1932, max: 2021 \\ 
		Gender & 41.49\% male, 58.51\% female  \\
		Distribution by race & 68\% Cau., 22\% Afr. Am., 1\% As., 9\% other  \\
		No. of diagnosis codes per patient & 14$\pm$11.1, min: 5, max: 121 \\
		No. of visits per patient & 9$\pm$6.6, min: 5, max: 63 \\
		\% of patients without labs & 14.33\%  \\
		\% of patients without procedures & 1.64\%  \\
		\% of patients without BMI & 21.74\%  \\ 
		\% of patients without smoking status & 27.11\% \\
		\% of deceased patients & 14.52\% \\  
		\bottomrule
	\end{tabular}
\end{table}
\subsection{Fine-Tuning Cohorts}  \label{ssec:cancer_cohort}
To validate the model's performance on cancer-specific tasks, we limited patients to have at least five diagnoses, regardless of the number of visits, in order to incorporate enough information for valid predictions. At least one of these diagnoses must be a cancer code (ICD-10 C[0-99]). This cohort consists of 437,903 cancer patients (31.67\% deceased within 6 months and 38.45\% within 12 months of first cancer diagnosis), split into three datasets (training (80\%), validation (10\%), test (10\%)), with each patient who is also part of the pre-training cohort described in section \ref{ssec:cohort} being assigned to the same data split as in the other cohort to avoid data leakage.\newline
We constructed the heart failure readmission cohort similarly, but did not restrict patients to a specific number of visits or diagnoses as long as one code was a heart failure code (ICD-10 I50). Again, we felt this was an appropriate use-case for clinicians. This resulted in a cohort of 503,161 heart failure patients (28.24\% readmitted within 30 days), split into three data sets. The detailed statistics of these two cohorts can be found in the appendix \ref{app:ft_cohorts}.

\subsection{Data Processing} 
For the diagnoses, we mapped all ICD-9 codes to ICD-10 codes according to the general equivalence mappings provided by the National Bureau of Economic Research\footnote{\url{https://www.nber.org/research/data/icd-9-cm-and-icd-10-cm-and-icd-10-pcs-crosswalk-or-general-equivalence-mappings}}. Furthermore, only primary diagnoses are considered, as we wanted to focus on the most important diagnostics. Similar to \cite{Meng2021} and \cite{Choi2015}, we limited the diagnosis codes to three characters to maintain a reasonable amount of relevant detail. Per visit diagnoses were de-duplicated to avoid biasing the model towards recurring codes during long visits. After de-duplication, patients with more than 128 diagnoses were discarded, as only 0.625\% of all patients had more than 128 diagnosis codes.
In addition, we included procedures\footnote{Optum\textsuperscript{\tiny\textregistered} EHR defines procedures as medications as well as disease screenings, surgical procedures and other medical services.} and laboratory types\footnote{Urinalysis, haematology, special chemistry, special laboratory, chemistry and blood gas} in the model. As with diagnoses, procedure codes and laboratory types were de-duplicated per visit.\newline
In addition to diagnoses, procedures, laboratory types, age and gender, we incorporated two of the least sparse observations in the dataset, which are empirically known to have high predictive power for disease trajectories: BMI and smoking status. If a value is missing for a particular visit, the last previous value is taken. Other observations with higher coverage such as pulse, body temperature and blood pressure were not taken into account, as they usually have a more granular influence on diseases. 

\section{Results} 

\subsection{General Evaluation Approach} 
Besides different baseline transformer models (BEHRT and Med-BERT), we also trained a version of ExBEHRT on the two pre-training objectives (MLM + PLOS) proposed in the publication of Med-BERT (\cite{Rasmy2021}) to verify potential benefits of having a second pre-training objectives (referred to as ExBEHRT+P). 

\subsection{Pre-Training Results} 
For pre-training ExBEHRT, we used the hyperparameters and model architecture proposed by the BEHRT paper. To ensure a fair comparison, we used the same amount of attention layers (6) and heads (12) as well as embedding dimension (288) for all three BERT-based models. For BEHRT and Med-BERT, we used the implementation provided by the authors of the corresponding publications\footnote{The source code can be found here: \url{https://github.com/deepmedicine/BEHRT} and \url{https://github.com/ZhiGroup/Med-BERT}}. All models were trained for 40 epochs on a Tesla T4 GPU with 16GB memory, where, as in the original publication, the epoch with the highest micro-averaged MLM precision score\footnote{Per default, the precision score is evaluated at a 0.5 threshold. Certainly, one could apply additional model calibration, but we hypothesized that this could introduce bias before fine-tuning.} was selected. We further report the balanced accuracy to get a better sense of the overall performance of the models.
\begin{table}[h!]
	\footnotesize
	\centering 
	\caption{Pre-Training results of various models.}
	\begin{tabular}{lllll}
		\toprule
		& \textbf{BEHRT} & \textbf{Med-BERT} & \textbf{ExBEHRT} & \textbf{ExBEHRT+P}\\
		\midrule
		Precision  & 54.6\% & 56.2\% & \textbf{64.2\%} & 63.9\% \\
		Balanced Accuracy  & 8.86\% & 8.03\% & \textbf{16.58\%} & 15.52\% \\ 
		\bottomrule
	\end{tabular}
	\label{tab:pretr_results} 
\end{table}\newline
\noindent As presented in table \ref{tab:pretr_results}, adding additional features to BEHRT significantly increases the pre-training performance in diagnosis code prediction. Adding a second pre-training objective slightly harms the MLM performance, but could nevertheless lead to improved fine-tuning performance due to additional context.

\subsection{Fine-Tuning on Event Prediction Results}
For this set of tasks we report the metrics commonly used to evaluate algorithms that perform binary predictions: area under the receiver operating characteristic curve (AUROC), average precision score (APS) and the precision at the 0.5 threshold. We used their micro-averaged implementations to follow the line of previous work and ensure a more robust assessment of the overall performance. More details on the hyperparameter optimization process can be found in appendix \ref{app:gridsearch}. We denote the task of prediction of death within N months after the first cancer prediction as \textit{Death in 6M} and \textit{Death in 12M} and predicting readmission of patients within 30 days of their first heart attack as \textit{HF readmit}.\newline
In addition to comparing the performance of our algorithm with that of BEHRT, we also benchmarked against two of the best performing "conventional" machine learning algorithms for tabular data, XGBoost (XGB, \cite{Chen2016}) and Logistic Regression (LR). Details on the preprocessing of the data and the tuning of the hyperparameters can be found in appendix \ref{app:xgb}.
\begin{table}[h!]
	\tiny
	\centering 
	\caption{Average fine-tuning results of various models and their standard deviations.}
	\begin{tabular}{llllllll}
		\toprule
		\textbf{Task} & \textbf{Metric} & \textbf{LR} & \textbf{XGB} & \textbf{BEHRT}& \textbf{Med-BERT}& \textbf{ExBEHRT} & \textbf{ExBEHRT+P}\\
		\midrule
		Death in 6M & \makecell{APS \\ AUROC \\ Precision } &\makecell{42.8$\pm$0.0\% \\ 63.5$\pm$0.0\% \\ 73.0$\pm$0.1\%} &\makecell{45.5$\pm$0.1\% \\ 66.4$\pm$0.1\% \\ 74.3$\pm$0.1\%} & \makecell{47.7$\pm$0.4\% \\ 66.7$\pm$0.6\% \\ 75.2$\pm$0.2\%} &\makecell{46.2$\pm$0.4\% \\ 65.3$\pm$0.3\% \\ 74.5$\pm$0.1\%} &  \makecell{ \textbf{53.1$\pm$0.3\%} \\ \textbf{71.5$\pm$0.5\%} \\ \textbf{78.1$\pm$0.1\%}} &  \makecell{52.6$\pm$0.3\% \\ 70.9$\pm$0.5\% \\ 77.9$\pm$0.1\%} \\ \midrule
		Death in 12M  & \makecell{APS \\ AUROC \\ Precision}&\makecell{51.6$\pm$0.0\% \\ 66.7$\pm$0.0\% \\ 70.4$\pm$0.1\%} &\makecell{45.5$\pm$0.1\% \\ 66.3$\pm$0.1\% \\ 74.4$\pm$0.1\%} &\makecell{55.5$\pm$0.1\% \\ 70.1$\pm$0.2\% \\ 73.2$\pm$0.1\%}  &\makecell{54.4$\pm$0.2\% \\ 68.9$\pm$0.3\% \\ 72.4$\pm$0.1\%} & \makecell{\textbf{59.8$\pm$0.2\%} \\ \textbf{74.3$\pm$0.4\%} \\ \textbf{76.4$\pm$0.1\%} } & \makecell{59.6$\pm$0.2\% \\ 73.8$\pm$0.4\% \\ 76.3$\pm$0.1\%}\\ \midrule
		HF readmit & \makecell{APS \\ AUROC \\ Precision}&\makecell{29.8$\pm$0.0\% \\ 51.9$\pm$0.1\% \\ 72.0$\pm$0.0\%} &\makecell{\textbf{31.3$\pm$0.1\%} \\ 53.6$\pm$0.1\% \\ 72.3$\pm$0.1\%} &\makecell{19.9$\pm$0.1\% \\ 51.2$\pm$0.1\% \\ 81.0$\pm$0.1\%} &\makecell{19.8$\pm$0.1\% \\ 51.0$\pm$0.1\% \\ 81.0$\pm$0.0\%}  & \makecell{30.0$\pm$1.6\% \\ 56.7$\pm$1.7\% \\ 78.7$\pm$0.2\%} & \makecell{25.1$\pm$0.1\% \\ \textbf{56.8$\pm$0.2\%} \\ \textbf{81.6$\pm$0.1\%}}\\
		\bottomrule
	\end{tabular}
	\label{tab:cancer_results} 
\end{table}\newpage
\noindent As shown in table \ref{tab:cancer_results}, the variants of ExBEHRT outperform the four baselines we created on all tasks. We also found that the addition of the second pre-training target PLOS can lead to slightly better performance in some scenarios, but is not superior overall. Nonetheless, XGBoost provides a higher APS than the transformer-based models on HF Readmit, but performs worse on the other metrics. \newline
For the two cancer mortality tasks, we further performed a cancer-specific evaluation of the ten most common cancers, as the different cancer types differ drastically in their expected outcomes. This evaluation can be found in appendix \ref{app:cancereval}. For all tasks presented here, we also examined the effects of omitting certain concepts to measure the impact of the new features. These ablations can be found in appendix \ref{app:featureablations}. In addition, we examined the effects on the positional variance of concepts within a visit. Since the model was trained to predict diagnosis codes at specific time points and not within a visit, there could be a possible bias that the model performs worse when the different features are mixed within a visit. The model should perform similarly regardless of which slot procedures and laboratory values are added to, as there is no temporal order within a visit. This ablation can be found in appendix \ref{app:shuffleablations}.
\subsection{Interpretability on Event Prediction Results}
For all interpretability experiments, we used the ExBEHRT model fine-tuned on the task \textit{Death in 6M}, meaning whether a cancer patient will decease within six months after their first cancer diagnosis. We visualize the interpretability for individual patients only, as both interpretability approaches presented here are example-based and not model-agnostic.
\subsubsection{Self-Attention Visualization}
\begin{figure}[h!]
	\centering 
	\includegraphics[width=10.6cm]{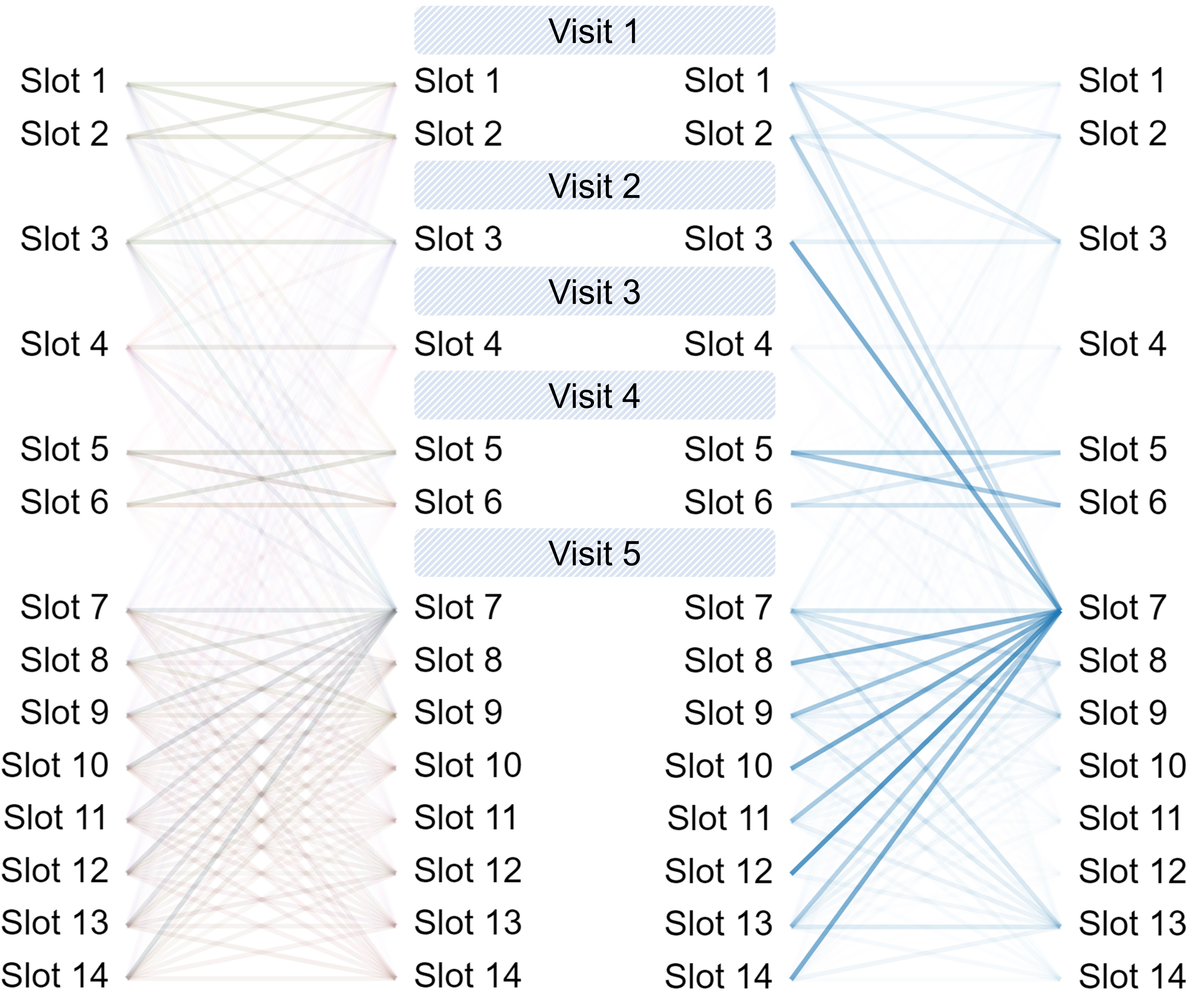} 
	\caption{Left: The self-attention of all 12 attention heads of the last layer of ExBEHRT. Higher opacity corresponds to higher attention. Right: The self-attention of one attention head of the last layer.}
	\label{fig:attention} 
\end{figure}

Analogous to previous work (\cite{Li2020}, \cite{Rasmy2021}, \cite{Meng2021}), we visualised the attention of the last network layer using BertViz (\cite{Vig2019}). However, since in all of these models the embeddings are summed before being passed through the network, self-attention has no way of attributing individual input features to the outcome. Nevertheless, we can draw conclusions about how the different slots interact with each other and which connections the model considers important. Figure \ref{fig:attention} shows the self-attention of a single patient in the last layer of ExBEHRT. The journey represents a 69-year-old woman who never smoked and died a year after being diagnosed with lung cancer. The left figure shows the attention of all 12 attention heads in this layer, while the right figure shows the attention of one single head. As expected, the model focuses heavily on the slots within a visit, as these slots are highly interconnected by definition. Although the model was not specifically trained on cancer codes, it pays close attention to slot 7 (slot containing the cancer diagnosis), suggesting that it has learned some correlation between the cancer diagnosis and the predicted outcome. Interestingly, slot 7 receives a lot of attention on the first and second visits, but not on the other two previous visits, suggesting that the model is able to learn causality over long periods of time. Table \ref{tab:slotplan} in appendix \ref{app:interpretability} contains information about all diagnoses, procedures and labs for each slot of the patient from the figure \ref{fig:attention}.
\newline
\subsubsection{Expected Gradients Interpretability}
\begin{figure}[h!]
	\centering 
	\includegraphics[width=11.5cm]{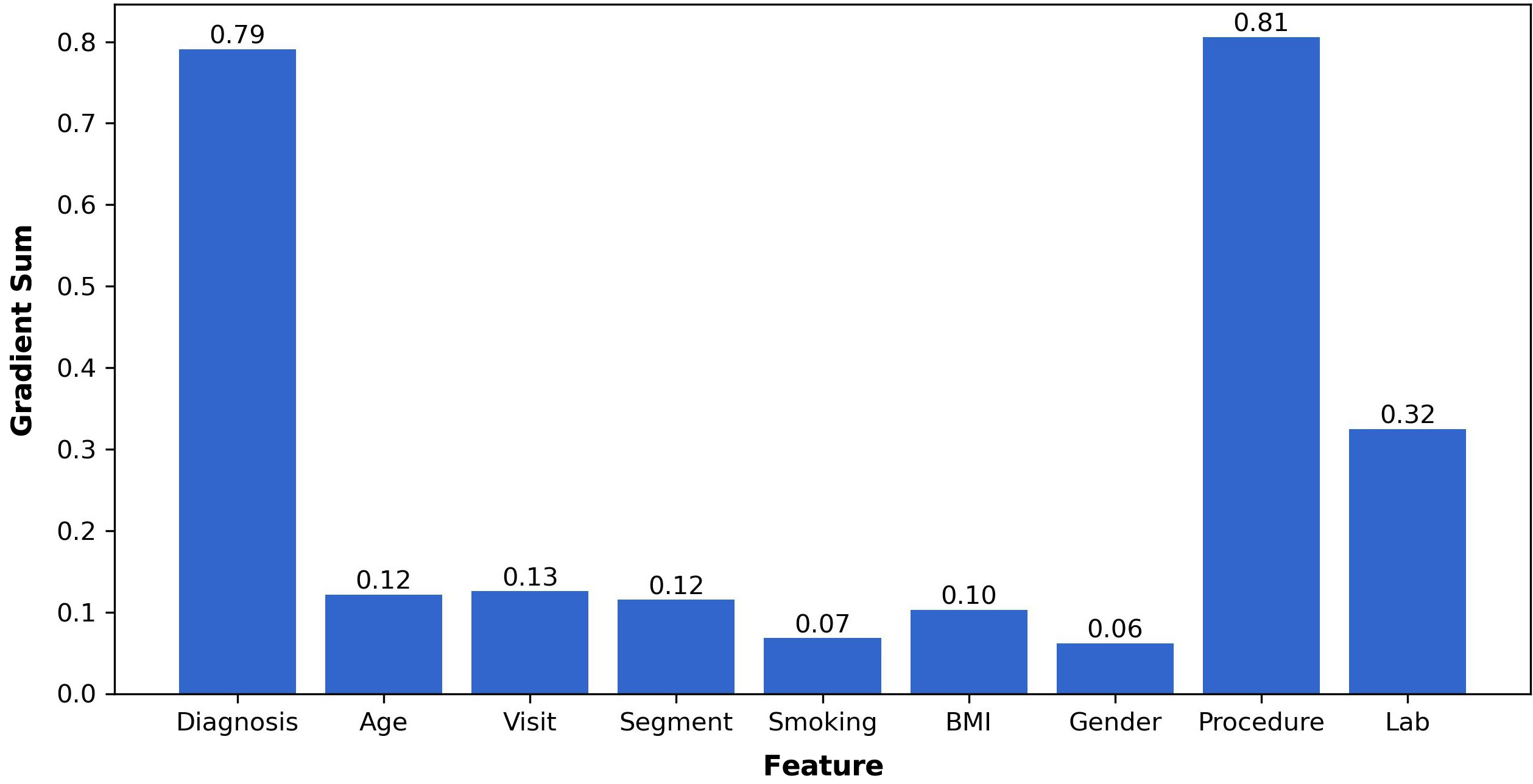} 
	\caption{The absolute sums of the expected gradients summed by input feature.}
	\label{fig:features} 
\end{figure}

\begin{figure}[h!]
	\centering 
	\includegraphics[width=11.5cm]{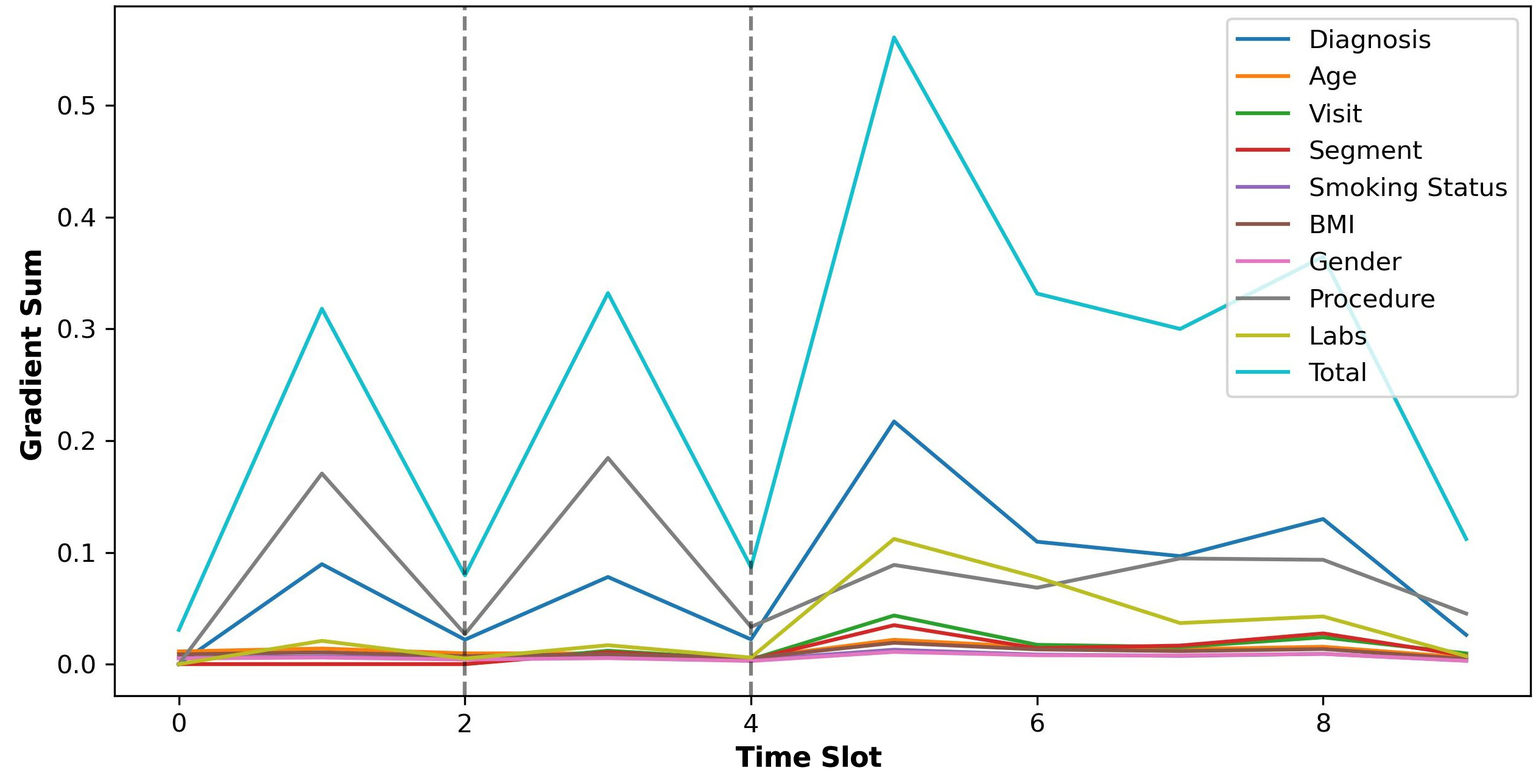} 
	\caption{The absolute sums of the expected gradients summed by input feature and time slot. The dotted lines indicate the next visit.}
	\label{fig:timely_sums} 
\end{figure}

\begin{figure}[h!]
	\centering 
	\includegraphics[width=15cm]{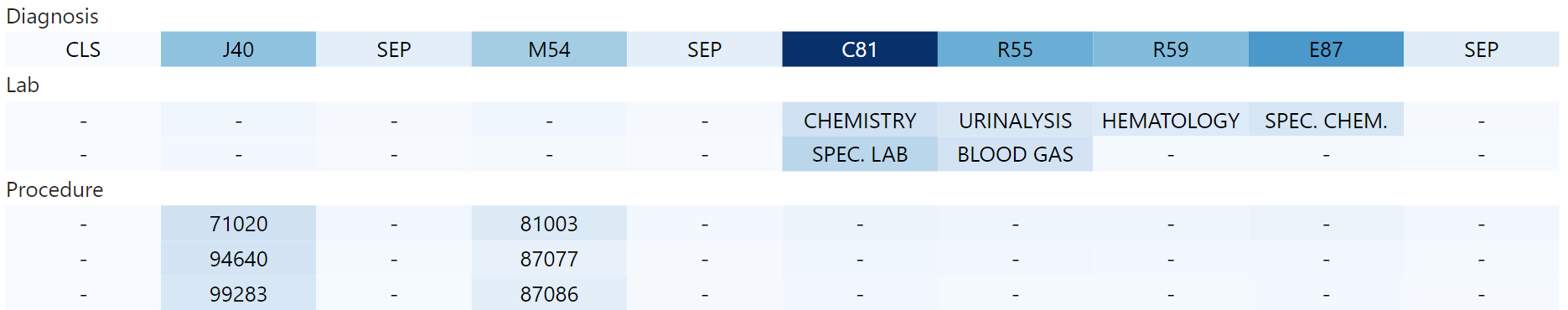} 
	\caption{A visualization of the absolute sums of the expected gradients of diagnoses, labs and procedures on a concept level. Darker colours represent higher values and the SEP tokens indicate the separation between two visits.}
	\label{fig:concepts} 
\end{figure}

Due to the limitations of self-attention visualisation, we have explored the technique Expected Gradients \citep{Erion2020} for more detailed interpretability. With this algorithm, we can infer the meaning of individual input tokens, which is not possible with self-attention. Since each token (diagnosis code, procedure code, age, etc.) is mapped to a 288-dimensional embedding before being passed to the model, we first calculated the expected gradients for the embedding and then summed the absolute values to obtain a single gradient value for each token. In this way, each individual token has an associated gradient that is linked to the output of the model and provides detailed insights into which medical concept has what impact on the prediction of the model. Our example patient is a 58-year-old woman who was a regular smoker. She died at the age of 65, three months after her blood cancer diagnosis. In figure \ref{fig:features}, we summed all expected gradients for each of the input features. This way, we can evaluate the feature importances on the output for a specific patient. For this patient, diagnoses and procedures (treatments \& medications) were by far the most importance features. With this visualization, we can further evaluate basic biases. For example, gender was not considered to be an important feature, indicating that predictions would be similar for a person with another gender. \newpage
\noindent In figure \ref{fig:timely_sums}, we visualized the absolute expected gradients for each of the features and summed them at each time slot. This way, we can evaluate the different feature importances over time to get a notion of where the model puts emphasis on. Interestingly, the model put more importance on what kind of medications \& treatments that patient received in the first two visits, where as in the last visit (the visit in which the patient was diagnosed with blood cancer), it put more importance on diagnoses and labs. Generally, slot 5, where the cancer was diagnosed, was attributed with the highest importance.
\newline
Figure \ref{fig:concepts} displays the absolute sums of gradients of each individual input token, providing a detailed interpretation of which medical concept has had what impact on the models prediction. Unsurprisingly, the cancer code C81 has had the biggest impact on the outcome. However, earlier codes like J40 or 71020 also contribute to the models prediction, indicating that the model includes information from the whole patient journey into its predictions.
\subsection{Patient Clustering} 
\begin{figure}[h!]
	\centering 
	\centerline{\includegraphics[width=15.5cm]{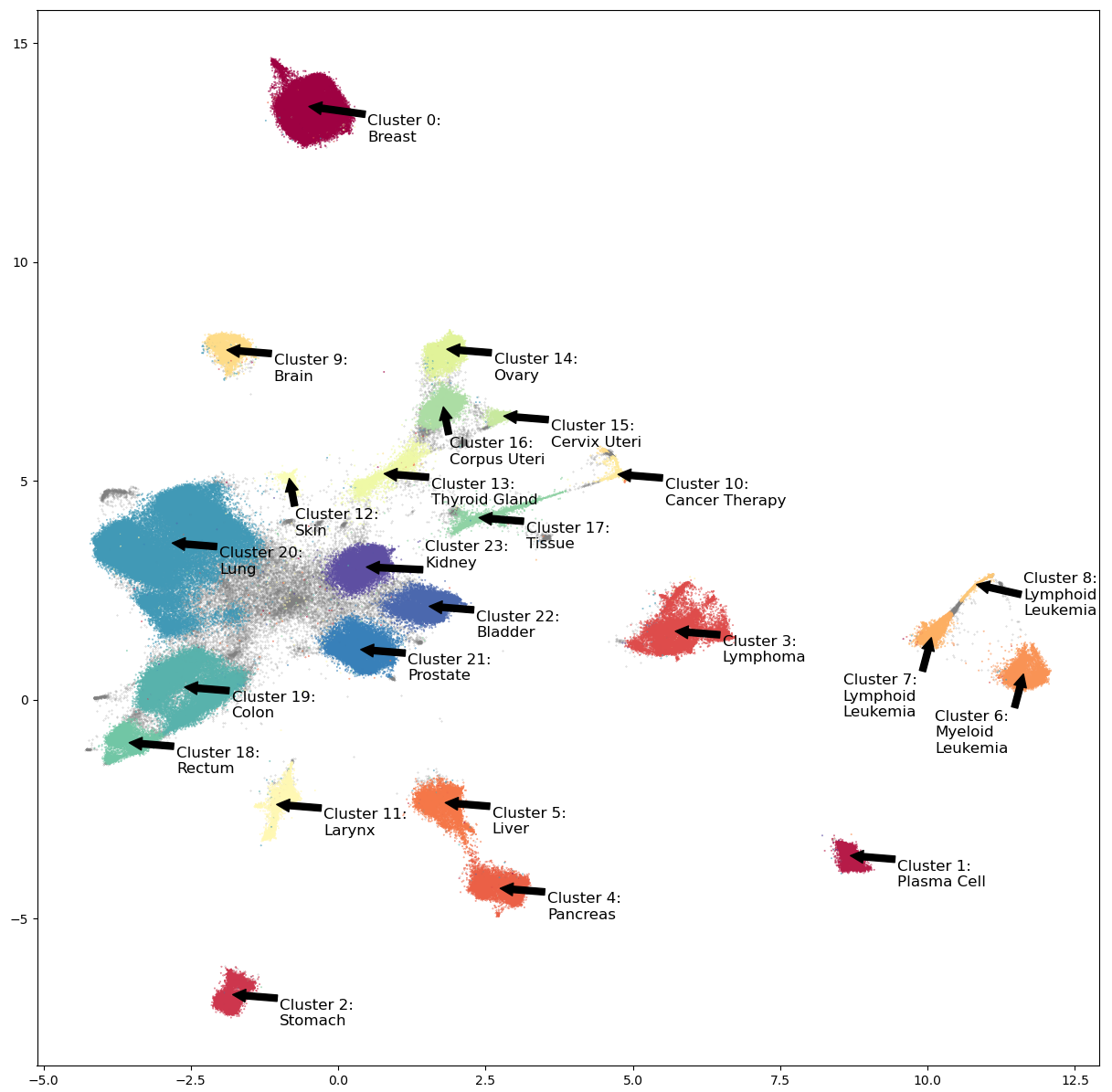} }
	\caption{The unsupervised cluster assignments from HDBSCAN, visualized with a 2-dimensional UMAP projection. The gray points are patients not assigned to any cluster (10\%). The labels indicate the most frequent diagnosis code of each cluster. Besides cluster 10, all labels are neoplasms.}
	\label{fig:clusters} 
\end{figure}

From the 260'645 cancer patients from the general cohort, HDBSCAN was able to cluster 90\% (234'575) into 24 clusters (mean: 9'774, min: 1'102, max: 47'722). As shown in figure \ref{fig:clusters}, the clusters are clearly separated spatially, indicating a distinct separation of the different cancer types. We labelled each cluster with the most occurring diagnosis code within this cluster, regardless of the type of code. Interestingly, similar concepts (e.g. cancer of female reproductive organs (clusters 14-16), different types of leukaemia (clusters 6-8)) or cancer of digestive organs (clusters 2, 4, 5, 18, 19) lay in areas close to each other, indicating a spatial logic within the disease types.
\newline
On average, the most common cancer diagnosis within a cluster was present in 84\% of patients assigned to that cluster, indicating a strong internal focus on cancer codes within the model. Of the 23 clusters, 22 had a unique cancer code as the most common diagnosis and included, on average, 85\% of all patients diagnosed with the corresponding cancer code. These two metrics indicate strong cross-cluster purity and homogeneity within the clusters. For a more detailed description of all clusters as well as the hyperparameters used in the different clustering steps, see appendix \ref{app:clustering} in table \ref{tb:cancer_clusters} and figures \ref{fig:most_common_1} and \ref{fig:most_common_2}.
\subsubsection{Disease Subtyping}
\begin{figure}[h!]
	\centering 
	\includegraphics[width=14.2cm]{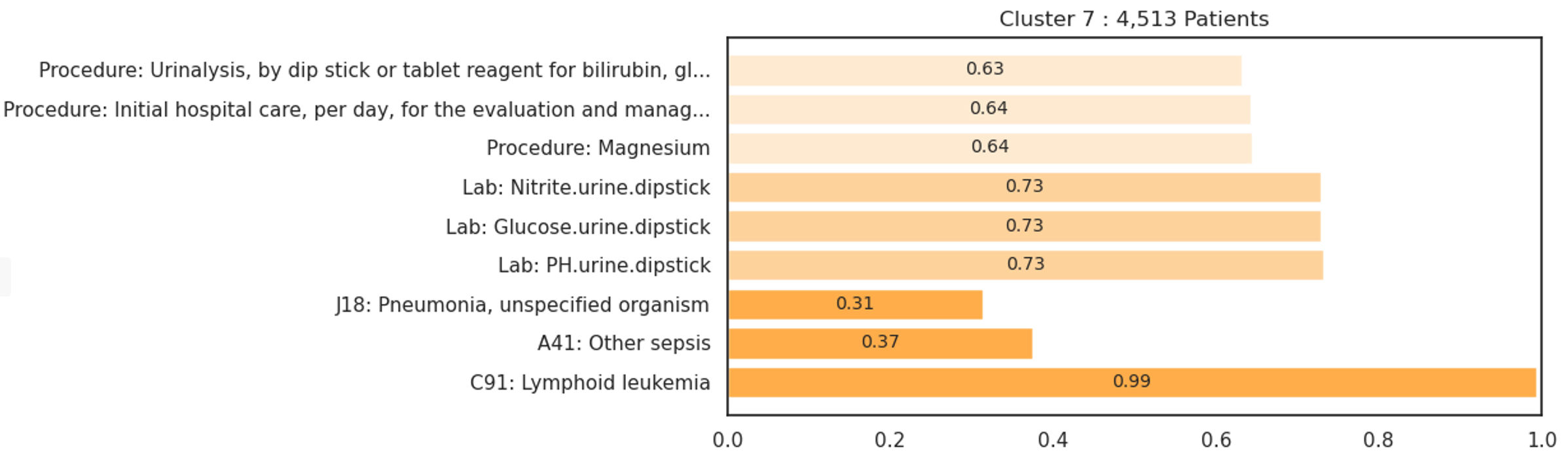} 
	\caption{The three most common procedures, labs and diagnoses for the CLL cluster.}
	\label{fig:cluster_7} 
\end{figure}

\begin{figure}[h!]
	\centering 
	\includegraphics[width=14.2cm]{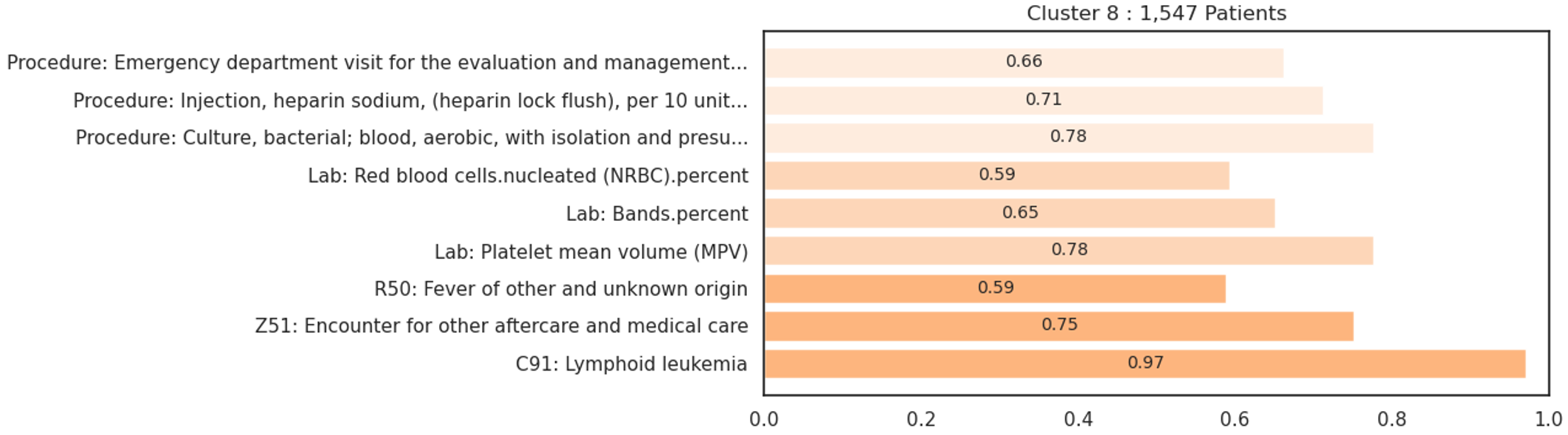} 
	\caption{The three most common procedures, labs and diagnoses for the ALL cluster.}
	\label{fig:cluster_8} 
\end{figure}

To draw conclusions about the internal clustering of HDBSCAN, we examined the most frequently occurring diagnoses, procedures and labs for each cluster. We focused only on concepts that occurred at least 5\% more frequently within the cluster than in the entire cohort. In this way, we ensured that very common diagnoses such as pelvic pain were not included in our cluster analysis. \newpage
\noindent A closer look at clusters 7 and 8 shows the potential of ExBEHRT to form subgroups of the same cancer type (Figures \ref{fig:cluster_7} \& \ref{fig:cluster_8}). Although almost all patients in both clusters have lymphocytic leukaemia, their diagnoses, procedures and applied laboratory tests differ considerably.
\newpage
\noindent Examination of the different patient characteristics (table \ref{tb:leuk_clusters}) of these two clusters (table \ref{tb:leuk_clusters}) shows that the model has indeed learned to distinguish between chronic lymphocytic leukaemia (CLL, cluster 7) and acute lymphocytic leukaemia (ALL, cluster 8) without having explicit information on these subtypes. As we limited the ICD-10 codes to three digits, only the general lymphocytic leukemia code C91 is given to the model without the subtypes C91.0 for ALL and C91.1 for CLL. In the table, \textit{\% of journey with cancer} indicates the ratio of the time between the first and last cancer diagnosis compared to the duration of the whole patient journey. \textit{Cancer-free} refers to the percentage of patients within a cluster, which have records of at least two visits after the last visit with a cancer diagnosis. The \textit{average death rate} comes directly from the Optum\textsuperscript{\tiny\textregistered} EHR database and unfortunately does not indicate the cause of death.
\begin{table}[h!]
	\scriptsize
	\centering
	\caption{Statistics of the two lymphoblastic leukemia clusters indicating a clear separation between CLL and ALL.}
	\begin{tabular}{ lll } 
		\hline
		\toprule
		\textbf{Metric} & \textbf{Cluster 7 (CLL)} & \textbf{Cluster 8 (ALL)}\\
		\midrule
		Median age & 70 & 5 \\
		Median birth year & 1946 & 2009 \\  
		Median BMI & 26 & 17 \\
		\% of men & 60.5\% & 55.2\% \\
		Average death rate & 54.7\% & 6.6\% \\
		\% of journey with cancer & 29.9\% & 45.5\% \\ 
		Cancer-free & 47.3\% &  48.1\%\\ 
		\bottomrule
	\end{tabular}
	\label{tb:leuk_clusters}
\end{table}\newpage
\noindent Another example, the pancreatic cancer cluster 4, shows that with a second pass of HDBSCAN on this cluster only, we can identify risk subgroups of pancreatic cancer. In all three identified clusters, more than 90\% of the patients actually do have pancreatic cancer and all share similar general characteristics.
\begin{figure}[h!]
	\centering 
	\includegraphics[width=7.5cm]{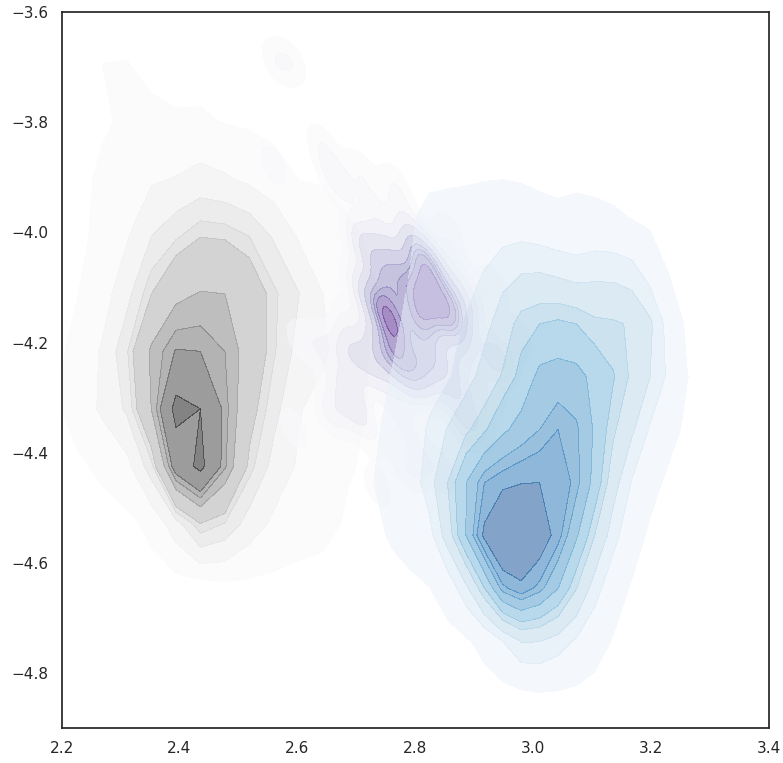} 
	\caption{The three identified patient subclusters with pancreatic cancer visualized with a kernel density estimate plot for visual clarity. Even though the three clusters generally share the same characteristics in diagnoses, Age, BMI etc., patients belonging to the smaller purple cluster died less frequently and recovered nearly twice as often from cancer compared to the other two clusters.}
	\label{fig:pancreas} 
\end{figure}\newline
\noindent However, as displayed in table \ref{tb:panc_clusters}, ExBEHRT identified one subgroup with significantly higher chance of recovering from cancer and a lower probability of death, even though this information was not provided to the model at any point.
\begin{table}[h!]
	\centering
	\caption{Statistics of the three pancreatic cancer clusters indicating a clear differentiation between higher risk (gray, blue) and lower risk patients (purple).}
	\begin{tabular}{ llll } 
		\hline
		\toprule
		\textbf{Metric} & \textbf{Gray} & \textbf{Blue} & \textbf{Purple}\\
		\midrule
		Median age & 67 & 68 & 68 \\
		Median birth year & 1950 & 1947 & 1944 \\  
		Median BMI & 25 & 25 & 26 \\
		\% of men & 52.3\% & 50.9\% & 60.0\% \\ 
		Average death rate & 76.5\% & 75.9\% & \textbf{70.0\%} \\
		\% of journey with cancer & 27.0\% & 24.0\% & \textbf{18.3\%} \\
		Cancer-free & 34.0\% & 36.9\% & \textbf{62.7\%} \\ 
		\bottomrule
	\end{tabular}
	\label{tb:panc_clusters}
\end{table}

\section{Discussion} 
In this study, we presented a novel method for adding patient features to BEHRT that significantly increases the predictive power for multiple downstream tasks in different disease domains. The novel method of stacking features vertically led to improvements in hardware requirements and benchmarks, and facilitates the possible extension to new concepts in the future. Given the large number and heterogeneity of patients with which the model was pre-trained, we are confident that ExBEHRT will generalise well to new data, patients and tasks. Combined with interpretability, the model offers more detailed insights into disease trajectories and subtypes of different patients than previous approaches, which could help clinicians form more detailed assessments of their patients' course and health. Furthermore, with a personalised understanding of patient groups, it is possible to identify unmet needs and improve patient outcomes.
\paragraph{Limitations and Future Work}\mbox{}\\
It is worth noting that the pre-training precision reported in BEHRT's original paper is higher than the one we were able to reproduce with the same model on our data (0.6597 (theirs) vs. 0.5456 (ours)). One possible explanation is that we drastically increased the task complexity since our model predicts a label out of 1916 instead of 300 diagnosis concepts. Nevertheless, we were able to show that additional features significantly improved both the quantitative and qualitative performance of the model, and we expect that this would also be the case when using the original dataset and the medical codes from BEHRT.\newline
Furthermore, it is extremely difficult to validate the quality, completeness and correctness of EHR datasets because EHR data is usually processed anonymously and comes from a variety of heterogeneous, fragmented sources. The pure nature of EHR data also introduces bias, as physicians may have an incentive to diagnose additional or other conditions, as medical billing is closely related to the number and type of diagnoses reported. \newline

\noindent In addition, there is also the question of bias and fairness in our results. In a possible next step, we would like to verify the results and interpretations of this work with clinicians to ensure robust and sound predictions given the interpretability we have acquired. In addition, we would like to test the generalisability of ExBEHRT to other clinical use-cases such as severity prediction and risk typing of other diseases and certain cancers. 


\bibliographystyle{plain}
\bibliography{library}
\newpage
\appendix
\section*{Appendix A. Fine-Tuning Cohorts} \label{app:ft_cohorts}

\begin{table}[h!]
	\centering
	\caption{Statistics of the cohort used for mortality prediction within six and twelve months after first cancer diagnosis.}
	\begin{tabular}{ ll } 
		\hline
		\toprule
		\textbf{Feature} & \textbf{Metric} \\
		\midrule
		Cohort Size & 437'903 \\
		Birth year &  1951$\pm$15, min: 1932, max: 2021 \\ 
		Gender & 50.42\% male, 49.58\% female  \\
		Distribution by race & 68\% Cau., 22\% Afr. Am., 1\% As., 9\% other  \\
		No. of diagnosis codes per patient & 15$\pm$12.6, min: 5, max: 118 \\
		No. of visits per patient & 7$\pm$6.1, min: 1, max: 63 \\
		\% of patients with outcome 6M & 31.67\% \\
		\% of patients with outcome 12M & 38.45\% \\
		\% of patients without labs & 18.81\%  \\
		\% of patients without procedures & 1.57\%  \\
		\% of patients without BMI & 21.83\%  \\ 
		\% of patients without smoking status & 29.11\% \\
		\% of deceased patients & 54.61\% \\  
		\bottomrule
	\end{tabular}
\label{tb:cancer_cohort}
\end{table}

\begin{table}[h!]
	\centering
	\caption{Statistics of the cohort used readmission prediction after heart failure within 30 days.}
	\begin{tabular}{ ll } 
		\hline
		\toprule
		\textbf{Feature} & \textbf{Metric} \\
		\midrule
		Cohort Size & 503'161 \\
		Birth year &  1945$\pm$14, min: 1932, max: 2021 \\ 
		Gender & 51.81\% male, 48.19\% female  \\
		Distribution by race & 77\% Cau., 16\% Afr. Am., 1\% As., 6\% other  \\
		No. of diagnosis codes per patient & 18$\pm$16.4, min: 1, max: 123 \\
		No. of visits per patient & 8$\pm$7.9, min: 1, max: 63 \\
		\% of patients with outcome & 28.24\% \\
		\% of patients without labs & 20.49\%  \\
		\% of patients without procedures & 2.56\%  \\
		\% of patients without BMI & 23.91\%  \\ 
		\% of patients without smoking status & 30.36\% \\
		\% of deceased patients & 51.63\% \\  
		\bottomrule
	\end{tabular}
\label{tb:hf_cohort}
\end{table}

\section*{Appendix B. Fine-Tuning Model Details} 

\subsection*{BEHRT Hyperparameter Selection} \label{app:gridsearch}
In order to tune the hyperparameters for downstream tasks, we applied a grid-search over the following parameters, selected the model with the best validation set performance and only then reported the performance on the test set:
\begin{itemize}
	\item Learning rate: $3e-5$, $4e-5$, $5e-5$
	\item Loss function: Cross-entropy, focal loss ($\gamma=2$), focal loss ($\gamma=5$), focal loss ($\gamma=2$, $\alpha=0.75$)
	\item Optimizer warmup: No warmup, warmup (warmup proportion=0.1, weight decay=0.01)
\end{itemize}

\begin{table}[h!]
	\centering 
	\caption{Fine-tuning hyperparameters on event prediction results of the BERT-based models.}
	\begin{tabular}{llllll}
		\toprule
		&  & \textbf{BEHRT}& \textbf{Med-BERT}& \textbf{ExBEHRT} & \textbf{ExBEHRT+P}\\
		\midrule
		Death in 6M & \makecell{LR \\ Loss \\ Warmup }  & \makecell{$5e-5$ \\ focal ($\gamma=2$) \\ yes }  & \makecell{$5e-5$ \\ focal ($\gamma=2$) \\ no } &  \makecell{$5e-5$ \\ focal ($\gamma=5$) \\ no } &  \makecell{$5e-5$ \\ focal ($\gamma=5$) \\ yes } \\ \midrule
		Death in 12M & \makecell{LR \\ Loss \\ Warmup }  & \makecell{$5e-5$ \\ focal ($\gamma=2$) \\ yes }  &  \makecell{$5e-5$ \\ focal ($\gamma=5$) \\ no } & \makecell{$5e-5$ \\ focal ($\gamma=5$) \\ no } & \makecell{$5e-5$ \\ focal ($\gamma=5$) \\ no }\\ \midrule
		HF readmit & \makecell{LR \\ Loss \\ Warmup }  & \makecell{$5e-5$ \\ focal ($\gamma=5$) \\ yes }  &  \makecell{$5e-5$ \\ focal ($\gamma=5$) \\ no } & \makecell{$5e-5$ \\ cross-entropy \\ yes } & \makecell{$5e-5$ \\ focal ($\gamma=5$) \\ yes }\\
		\bottomrule
	\end{tabular}
\end{table}

\subsection*{Conventional Models Data Pre-Processing \& Hyperparameter Selection} \label{app:xgb}
To establish a baseline for the event prediction tasks, we conducted experiments with XGBoost and Logistic Regression, two versatile machine learning algorithms for tabular data. For data pre-processing, we ensured that the models had the same information as the transformer-based approaches. For the cancer patient mortality prediction, this means the entire journey to the first cancer prediction, and for the readmission tasks, a time window of one year to the first heart failure. Since neither model can work with time-series data, we converted the patient journeys into multi-hot encoded features, with each column indicating whether and how often the patient was assigned a particular diagnosis, procedure or laboratory. We also included a one-hot-coded vector for smoking status, a column for number of visits, gender, and a patient's last and first BMI value and age. In this way, we wanted to ensure a fair comparison, as the models can access the same set of variables as ExBEHRT.\newpage
\noindent To tune the hyperparameters, we used the library Optuna (\cite{Akiba2019}), a renowned hyperparameter optimization framework and optimized the mean-squared error while optimizing the following parameters for XGB: learning rate (LR), maximum tree depth (max\_depth), number of estimators (n\_estimators), column sampling by tree (colsample), row subsampling (subsample) and the regulation parameter $\alpha$.
\begin{table}[h!]
	\centering 
	\caption{The best XGBoost hyperparameters found by Optuna for the fine-tuning prediction tasks.}	\begin{tabular}{lcccccc}
		\toprule
		& \textbf{LR} & \textbf{max\_depth}& \textbf{n\_estimators} & \textbf{colsample} & \textbf{subsample} & \textbf{$\alpha$}\\
		\midrule
		Death in 6M & 0.0729 & 12 & 538 & 0.8095 & 0.8119 & 0.1989 \\
		Death in 12M & 0.1180 & 8 & 666 & 0.3705 & 0.7091 & 0.7988 \\
		HF readmit & 0.0127 & 19 & 1148 & 0.4438 & 0.9937 & 0.5952 \\
		\bottomrule
	\end{tabular}
\end{table}\newline
We conducted a similar hyperparameter search for logistic regression with the following parameters: class weight (balanced or none), the equation solver (Newton-Cholesky, Sag or Saga), and the two regularization parameters penalty and C.
\begin{table}[h!]
	\centering 
	\caption{The best Logistic Regression hyperparameters found by Optuna for the fine-tuning prediction tasks.}	\begin{tabular}{lcccc}
		\toprule
		& \textbf{Class weight} & \textbf{Solver}& \textbf{Penalty} & \textbf{C}\\
		\midrule
		Death in 6M & None & Sag & L2 & 186.8192 \\
		Death in 12M & None & Saga & None & 141.8034 \\
		HF readmit & None & Newton-Cholesky & None & 162.0031\\
		\bottomrule
	\end{tabular}
\end{table}

\section*{Appendix C. Fine-Tuning Cancer-Wise Evaluation} \label{app:cancereval} 
Since different cancer types empirically have highly varying predicted outcomes, we evaluated the model performance on the ten most common cancer types in the test dataset to spot potential bias.
\begin{table}[h!]
	\tiny
	\centering
	\caption{The AUROC scores of ExBEHRT for the ten most common cancer types in the test dataset.}
	\begin{tabular}{lllllll}
		\toprule
		\textbf{ICD-10} & \textbf{\makecell{\#Patients \\ with code }} & \textbf{\makecell{Cancer \\ description }}                             & \textbf{\makecell{\%Patients with \\ outcome 6M }} & \textbf{\makecell{ROCAUC \\ Death in 6M }} & \textbf{\makecell{\%Patients with \\ outcome 12M }} & \textbf{\makecell{ROCAUC \\ Death in 12M }} \\
		\midrule
		C34                  & 7312                          & Bronchus and lung                 & 44.46\%                             & 0.7186                      & 52.38\%                              & 0.7172                       \\
		C18                  & 3710                          & Colon                             & 21.4\%                              & 0.6652                      & 26.95\%                              & 0.6960                       \\
		C50                  & 2924                          &Breast                            & 20.11\%                             & 0.7031                      & 24.66\%                              & 0.7257                       \\
		C61                  & 2057                          & Prostate                          & 17.36\%                             & 0.7255                      & 21.78\%                              & 0.785                        \\
		C25                  & 1986                          & Pancreas                          & 48.59\%                             & 0.7058                      & 58.51\%                              & 0.6873                       \\
		C80                  & 1965                          & No specification of site        & 54.4\%                              & 0.7064                      & 61.58\%                              & 0.6840                       \\
		C64                  & 1880                          & Kidney, except renal pelvis       & 16.76\%                             & 0.6824                      & 20.96\%                              & 0.7336                       \\
		C71                  & 1760                          & Brain                             & 30.06\%                             & 0.6789                      & 40.06\%                              & 0.7111                       \\
		C22                  & 1594                          & Liver & 49.18\%                             & 0.6989                      & 54.83\%                              & 0.7123                       \\
		C67                  & 1472                          & Bladder                           & 27.17\%                             & 0.6775                      & 33.97\%                              & 0.7117       \\
      	\bottomrule       
	\end{tabular}
\end{table}\newline
Even tough labels are heavily imbalanced for some cancer types and frequencies vary drastically, the model is able to perform well on the individual cancer subgroups and we couldn't identify a bias towards certain subgroups.
\newpage
\section*{Appendix D. Ablations} 
\subsection*{Feature Ablations}\label{app:featureablations}
Here, we conducted ablations on which features independently yields the biggest improvement on the three proposed fine-tuning tasks. Generally, adding procedures yields the most significant performance boost. Nevertheless, combining all features yields to the best performance for all but one metric in one task (Precision for \textit{HF readmit}).
\begin{table}[h!]
	\scriptsize
	\centering 
	\caption{Ablations on fine-tuning on event prediction results of our ExBEHRT model trained with different features starting with BEHRT (Diagnosis + Age + Segment + Visit), adding only procedures, only labs and only observations (BMI, smoking status, gender).}
	\begin{tabular}[t]{lllllll}
		\toprule
		\textbf{Task} & \textbf{Metric} & \textbf{BEHRT}& \textbf{\makecell{BEHRT \\ + Procedures }}& \textbf{\makecell{BEHRT \\ + Labs }} & \textbf{\makecell{BEHRT \\ + Observations }} & \textbf{ExBEHRT}\\
		\midrule
		Death in 6M & \makecell{APS \\ AUROC \\ Precision }  & \makecell{0.4778 \\ 0.6674 \\ 0.7520} &\makecell{0.5206 \\ 0.7034 \\ 0.7762}&\makecell{0.4786 \\ 0.6696 \\ 0.7527}&\makecell{0.4871 \\ 0.6791 \\ 0.7566} &  \makecell{ 0.5312 \\ 0.7154 \\ 0.7811}  \\ \midrule
		Death in 12M  & \makecell{APS \\ AUROC \\ Precision} &\makecell{0.5545 \\ 0.7010 \\ 0.7319}  &\makecell{0.5938 \\ 0.7352 \\ 0.7611}&\makecell{0.5566 \\ 0.7040 \\ 0.7333}&\makecell{0.5609 \\ 0.7063 \\ 0.7369} & \makecell{0.5978 \\ 0.7432 \\ 0.7639 } \\ \midrule
		HF readmit & \makecell{APS \\ AUROC \\ Precision} &\makecell{0.1994 \\ 0.5117 \\ 0.8102} &\makecell{0.2412 \\ 0.5582 \\ 0.8145}&\makecell{0.1998 \\ 0.5117 \\ 0.8106}&\makecell{0.1994 \\ 0.5115 \\ 0.8103}  & \makecell{0.2966 \\ 0.5677 \\ 0.7873} \\
		\bottomrule
	\end{tabular}
\end{table}

\subsection*{Visit Shuffle Ablations}\label{app:shuffleablations}
For these ablations, we randomly shuffled the diagnoses, procedures and labs within each visit of 50 random patients from the test dataset a total of 50 times. All other characteristics are time-dependent, as they also show changes within a visit, and were therefore excluded from shuffling. We followed the same principle for all three fine-tuning tasks to check for positional invariance within each visit. To determine the label purity for a patient, we divided the amount of the most frequently predicted label of \{0.1\} by 50. A value of 1.0 would mean that the model predicted the same label on every shuffle, indicating positional invariance. We then took the mean of all 50 patients to determine the predicted label purity.
\begin{table}[h!]
	\centering 
	\caption{The predicted label purity of the best-performing ExBEHRT models on all three subtasks.}	
	\begin{tabular}{lc}
		\toprule
		\textbf{Task} & \textbf{Aggregated Label Purity}\\
		\midrule
		Death in 6M & 0.9820$\pm$0.05  \\
		Death in 12M & 0.9884$\pm$0.06  \\
		HF readmit & 0.9910$\pm$0.07 \\
		\bottomrule
	\end{tabular}
	\label{tab:purity}
\end{table}\newline
As shown in table \ref{tab:purity}, the purity on each task is very close to 1.0, indicating that the model is very robust regarding the in-visit order of diagnoses, procedures and labs and doesn't put significant focus on it.  

\section*{Appendix E. Clustering Details} \label{app:clustering}

\subsection*{Clustering Hyperparameters}
The hyperparameters for the procedure of the clustering from section \ref{ssec:cancer_clust} were set as follows:
\begin{itemize}
	\item UMAP for 10D representations (step $b$):
	\begin{itemize}
		\item Size of local neighbourhood: 100
		\item Minimum distance apart : 0.0
		\item Distance metric: euclidean 
	\end{itemize}
	\item HDBSCAN clustering of the 10D representations (step $c$):
	\begin{itemize}
		\item Minimum cluster size: 1000
		\item Minimum samples: 1000
	\end{itemize}
	\item UMAP for 2D representations (step $d$): 
	\begin{itemize}
		\item Size of local neighbourhood: 100
		\item Minimum distance apart: 0.0
		\item Distance metric: euclidean 
	\end{itemize}
\end{itemize}
\newpage
\subsection*{Clustering Results}
\begin{table}[h!]
	\centering
	\caption{Statistics of the different clusters of cancer patients. The in-cluster percentage indicates how many patients within the cluster were diagnosed with the most occurring code of this cluster. The purity percentage indicates how many of the patients diagnosed with the most occurring code were actually assigned to the indicated cluster. Cluster 10 was excluded as it is not a cancer cluster.}
	\begin{tabular}{ lllll } 
		\hline
		\toprule
		\textbf{Cluster} & \textbf{Patients} & \textbf{Most occurring code} & \textbf{In-cluster} & \textbf{Purity}\\
		\midrule
		Cluster 0 & 22'053 & C50 & 98\% & 98\% \\
		Cluster 1 & 6'248 & C90 & 99\% & 94\% \\
		Cluster 2 & 8'250 & C16 & 58\% & 93\% \\
		Cluster 3 & 14'565 & C85 & 54\% & 89\% \\
		Cluster 4 & 10'642 & C25 & 95\% & 95\% \\
		Cluster 5 & 9'377 & C22 & 80\% & 88\% \\
		Cluster 6 & 7'263 & C92 & 88\% & 92\% \\
		Cluster 7 & 4'513 & C91 & 99\% & 67\% \\
		Cluster 8 & 1'547 & C91 & 97\% & 22\% \\
		Cluster 9 & 6'815 & C71 & 95\% & 83\% \\
		Cluster 11 & 7'399 & C32 & 32\% & 94\% \\
		Cluster 12 & 1'847 & C43 & 99\% & 89\% \\
		Cluster 13 & 5'458 & C73 & 66\% & 87\% \\
		Cluster 14 & 6'668 & C56 & 93\% & 92\% \\
		Cluster 15 & 2'174 & C53 & 99\% & 85\% \\
		Cluster 16 & 6'870 & C54 & 85\% & 90\% \\
		Cluster 17 & 3'705 & C49 & 70\% & 71\% \\
		Cluster 18 & 7'387 & C20 & 87\% & 89\% \\
		Cluster 19 & 20'772 & C18 & 82\% & 86\% \\
		Cluster 20 & 47'722 & C34 & 77\% & 93\% \\
		Cluster 21 & 13'119 & C61 & 93\% & 90\% \\
		Cluster 22 & 9'679 & C67 & 92\% & 90\% \\
		Cluster 23 & 9'400 & C64 & 93\% & 83\% \\ \midrule
		Total & 233'473 & - & 84\% & 85\% \\
		\bottomrule
	\end{tabular}
\label{tb:cancer_clusters}
\end{table}
\begin{figure}[hbt!]
	\centering
	\centerline{\includegraphics[width=15.2cm]{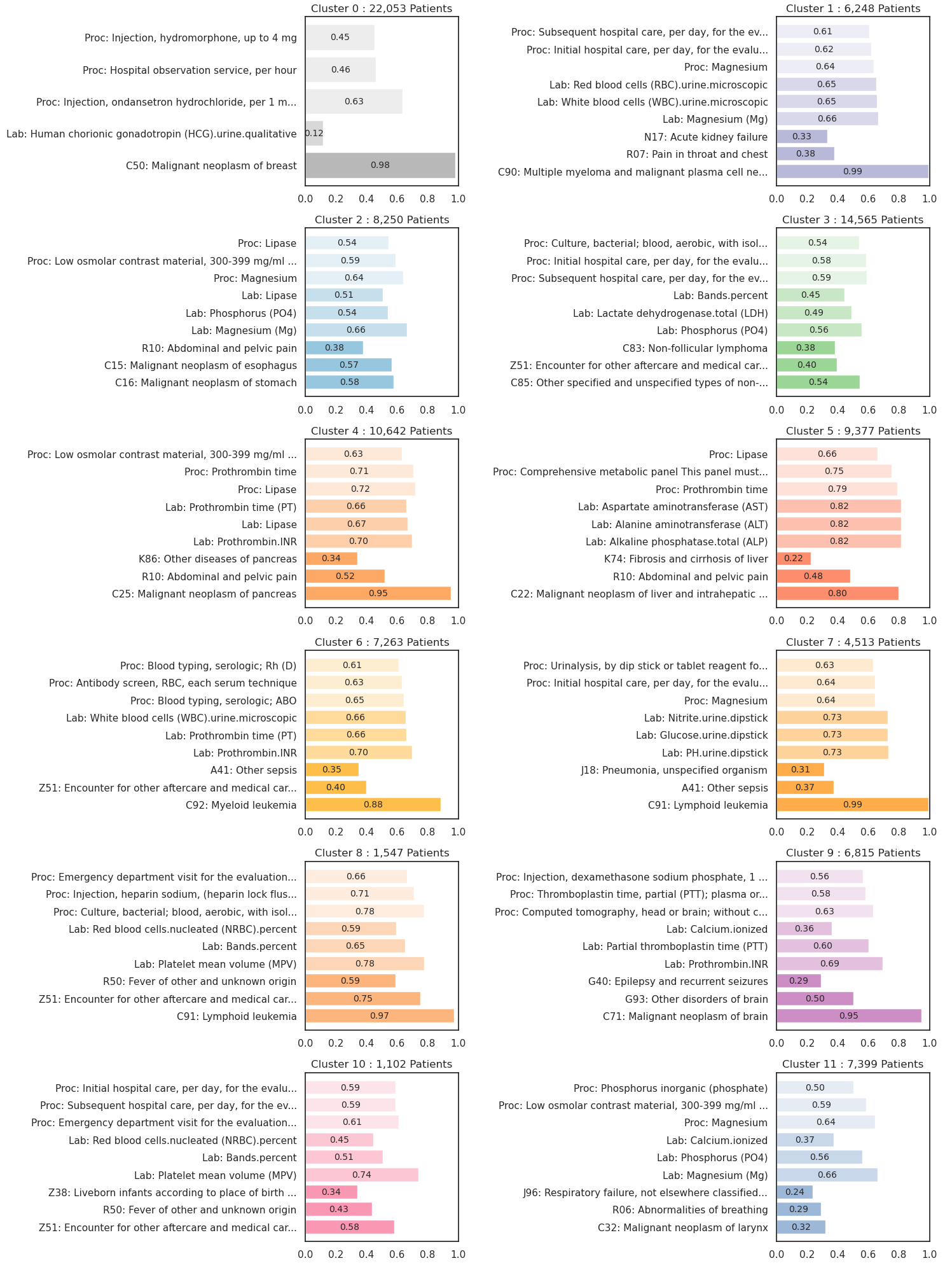} }
	\caption{The three most common procedures, labs and diagnoses for the clusters 0-11 identified by HDBSCAN. Each concept listed occurs within this cluster at least 5\% more often than in the overall cohort.}
	\label{fig:most_common_1} 
\end{figure} 
\begin{figure}[hbt!]
\centering
\centerline{\includegraphics[width=15.2cm]{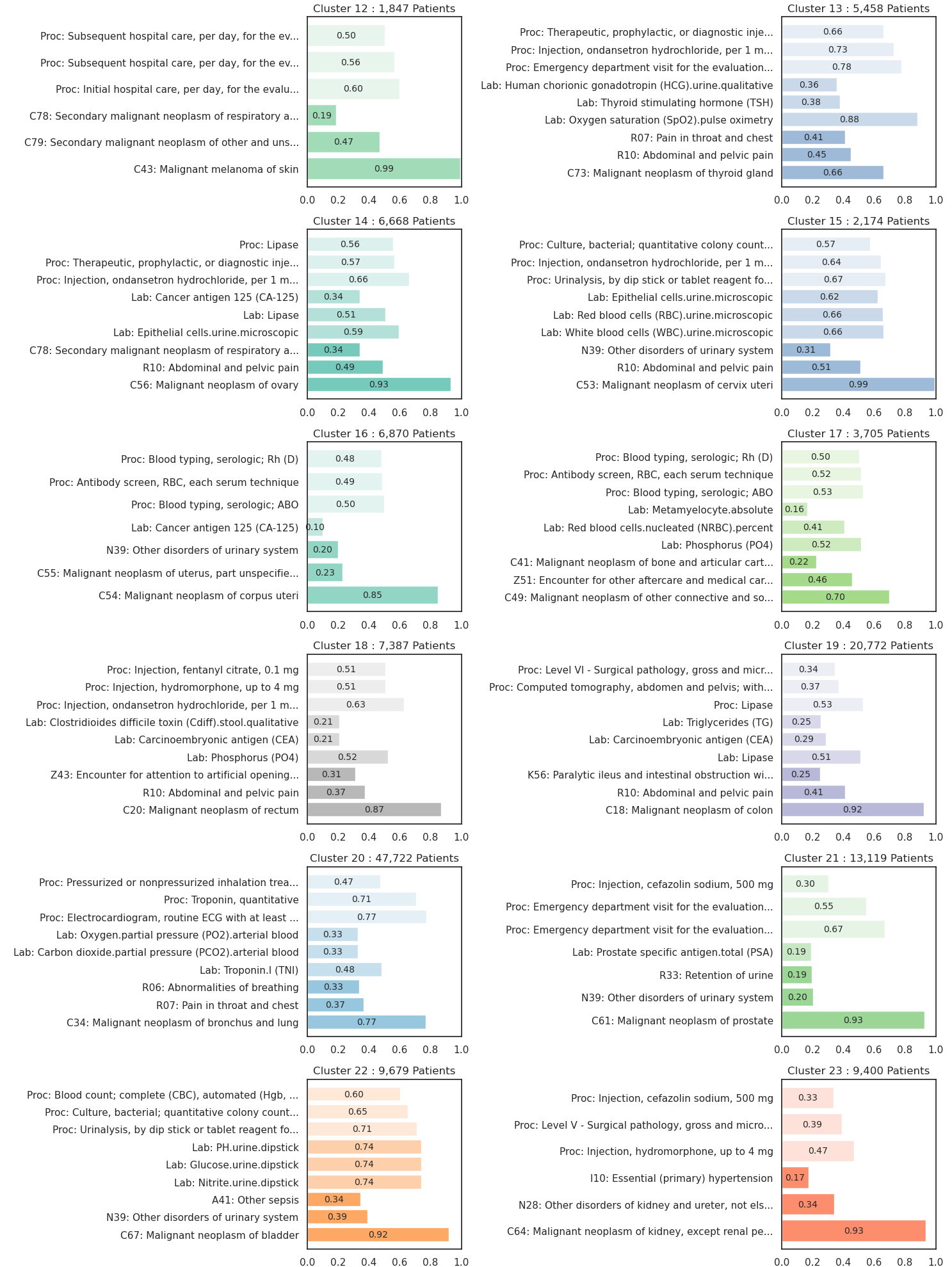} }
\caption{The three most common procedures, labs and diagnoses for the clusters 12-23 identified by HDBSCAN. Each concept listed occurs within this cluster at least 5\% more often than in the overall cohort.}
\label{fig:most_common_2} 
\end{figure}

\clearpage
\section*{Appendix F. Interpretability} \label{app:interpretability}
\begin{table}[hbt!]
	\centering 
	\caption{The corresponding diagnoses, procedures and labs for each slot of the patient used with the Self-Attention visualization methodology. The patient journey is cut at the first cancer diagnosis.}
	\begin{tabular}{llll}
		\toprule
		\textbf{Slot} & \textbf{Diagnosis} & \textbf{Procedures}                                                         & \textbf{Labs}         \\ \fatmidrule
		1             & E11                & 36415, 80076, 93041, 96374                                                  & Chemistry, urinalysis \\ \midrule
		2             & R55                & 80048, 85025, 94760, 99284                                                  & Hematology, blood gas \\ \fatmidrule
		3             & J40                & \makecell{36415, 71020, 80053, 83880, 85025, 93005, \\ 93041, 94760, 96360, 99284, A9270 }  & -                     \\ \fatmidrule
		4             & M17                & -                                                                           & -                     \\ \fatmidrule
		5             & S13                & 72100, 73520, 96372, J1885                                                  & Blood gas             \\ \midrule
		6             & M54                & 72125, 94760, 99285                                                         & -                     \\ \fatmidrule
		7             & C34                & 99223                                                                       & Chemistry             \\ \midrule
		8             & J18                & 93306                                                                       & Hematology            \\ \midrule
		9             & I46                & 93010                                                                       & Blood gas             \\ \midrule
		10            & R06                & 99232                                                                       & -                     \\ \midrule
		11            & J96                & -                                                                           & -                     \\ \midrule
		12            & I49                & -                                                                           & -                     \\ \midrule
		13            & J15                & -                                                                           & -                     \\ \midrule
		14            & R07                & -                                                                           & -                \\    
		\bottomrule        
	\end{tabular}
	\label{tab:slotplan} 
\end{table}
\end{document}